# Adaptive speed planning for Unmanned Vehicle Based on Deep Reinforcement Learning

Hao Liu [1],[*]
Northeastern University
Shenyang, China
liuhao@stumail.neu.edu.cn

Yi Shen[1]
University of Michigan, Ann Arbor
USA
shenrsc@umich.edu

Wenjing Zhou[1]
University of Michigan Ann Arbor
USA
wenjzh@umich.edu

Yuelin Zou[2]
Columbia University
USA
yz4198@columbia.edu

Chang Zhou[2]
Columbia University
USA
mmchang042929@gmail.com

Shuyao He[2]
Northeastern University
USA
he.shuyao@northeastern.edu

*Abstract*—In order to solve the problem of frequent deceleration of unmanned vehicles when approaching obstacles, this article uses a Deep Q-Network (DQN) and its extension, the Double Deep Q-Network (DDQN), to develop a local navigation system that adapts to obstacles while maintaining optimal speed planning. By integrating improved reward functions and obstacle angle determination methods, the system demonstrates significant enhancements in maneuvering capabilities without frequent decelerations. Experiments conducted in simulated environments with varying obstacle densities confirm the effectiveness of the proposed method in achieving more stable and efficient path planning.

## I. INTRODUCTION

In the development of future technologies, online path planning for unmanned vehicles is particularly critical, especially in complex urban transportation networks. In recent years, deep reinforcement learning(DRL) has become a cutting-edge technology to solve such problems. Li et al. [1] proposed a reinforcement learning-based routing algorithm in their research, specifically designed for large-scale street networks, demonstrating the effectiveness of DRL in handling large-scale and dynamic environments[2].

With the advancement of technology, security issues have become increasingly prominent[3]. Zhang et al. [4] detected Specter attacks through hardware-level machine learning methods, emphasizing the importance of hardware security when implementing complex algorithms[5].

In addition, the applications of DRL are not limited to the transportation field[6]. In the field of data processing, research by Li P et al. [7] shows that the advanced technology of dual-modal convolutional neural network and high-resolution semantic segmentation can effectively support complex data analysis[8].

In terms of other applications of deep learning, Yan[9] demonstrated research on using LSTM neural networks to predict oil field production and atmospheric electric field instruments to predict lightning locations. These studies not only enhanced the prediction capabilities of the model, but also expanded the use of deep learning in Applications in Environmental Science[10]. At the same time, research on hardware security and robot interaction [11] also shows that the simulation of human behavior and semi-supervised learning provide a new perspective for autonomous driving systems.

Some scholars proposed a new planning algorithm for trajectory tracking control of single-wheel mobile robots, demonstrating the possibility of precise control in complex environments. At the same time, Wang et al. [12] designed a dual-spring damping system for outdoor automatic guided vehicles (AGV). This physical improvement provides stability for the robot to operate on uneven ground[13].

Some people made a theoretical breakthrough in graph neural network training acceleration, which is particularly important for processing large amounts of traffic network data and making route decisions in real time[14].

Regarding hardware aspects, Wen et al. [15] examined the management of hardware memory within prospective mobile hybrid storage systems., which provides hardware support for processing large-scale data. Feng et al. [16] demonstrated a human tracking robot using ultra-wideband technology, which has potential application value in automatic obstacle avoidance and tracking of unmanned vehicles [17].

Finally, multimodal learning methods have also been widely studied [18]. Gao et al.[19] applied machine learning-based algorithm models and fuzzy controllers in the design of traffic sign recognition and automatic parking systems, demonstrating the wide application of machine learning in actual traffic applications. Liu et al. [20] used particle filters to perform simultaneous localization and map construction (SLAM) for vehicle positioning, which is crucial for the navigation of unmanned vehicles in unknown environments[21]. DRL technology has shown great potential in many fields [22] and is widely used in real life [23], such as data adaptation and parameter tuning [24] [25].

This article has made some improvements to the speed planning part of the unmanned vehicle. First, the vehicle speed is coupled with the angle between the vehicle and the obstacle, and the coupling relationship is integrated into the reward function. Secondly, the DDQN algorithm is used to replace the local path planning of the unmanned vehicle. module; finally, the test of vehicle speed planning under different environments was completed in the Gazebo simulation environment.

## II. MODEL BUILDING

### A. DQN algorithm

Deep Q-networks (DQN) combine Q-learning principles with deep neural networks to handle complex environments with high-dimensional state spaces. The DQN algorithm has revolutionized the field of reinforcement learning by employing deep neural networks to approximate optimal action value functions, which are critical for making informed decisions in a variety of states. The neural network is trained to minimize a loss function derived from the time difference error, combined with Q-learning updates and gradient descent optimization.

The core of the DQN is the update rule for the action-value function, which iteratively improves the policy. The DQN algorithm utilizes the Q-learning framework to derive an optimizable loss function for training the neural network. The update equation is defined as follows (1):

$$Q(S_t, A_t) \leftarrow Q(S_t, A_t) + \alpha [R_{t+1} + \gamma \max_{a'} Q(S_{t+1}, a') - Q(S_t, A_t)] \quad (1)$$

The training procedure involves reducing a loss function that quantifies the discrepancy between predicted and target Q-values. Per equation (1), the loss function for the DQN algorithm is articulated as follows (2):

$$L(\theta_t) = \mathbb{E}\left[\left(Y_t^{DDQN} - Q(S_t, A_t; \theta_t)\right)^2\right] \quad (2)$$

The target Q-value represents the anticipated future rewards, discounted by the factor $\gamma$, and it is crucial for the stability of the learning process:

$$\text{Target}Q = r + \gamma \max_{a'} Q(S_{t+1}, a'; \theta^-) \quad (3)$$

The DQN employs a separate target network to stabilize the learning updates. The parameters of the target network are periodically updated with the parameters from the primary network ($\theta$), preventing the rapid oscillation of target values that could destabilize the learning process.

### B. DDQN algorithm

Double Deep Q-Network (DDQN) enhances the original DQN by addressing the issue of Q-value overestimation. Both DQN and DDQN employ deep neural networks to approximate the Q-value function in environments with high-dimensional state spaces, marking a major advancement in reinforcement learning.

The DDQN algorithm represents an important iterative improvement over DQN. By separating the action selection process from the Q-value evaluation, DDQN reduces the overoptimistic value estimates that can arise in DQN. The introduction of this subtle yet impactful change has been shown to produce more stable and reliable learning, and it prevents the agent from overvaluing actions during policy development. This, in turn, often leads to better performance on various benchmark tasks in reinforcement learning.

### C. Improved reward function

In some existing reinforcement learning algorithms, the setting of the reward function is simple, which will cause the mobile robot to pay too much attention to obstacle avoidance when performing steering control and speed control. As a result, even when the deviation between the speed direction and the obstacle is large, the performance will still be greatly reduced. The speed is to ensure avoiding obstacles and is not flexible enough in the planning process. A similar reward function is shown in Equation (4):

$$R = \begin{cases} 10 & nocollision \\ -100 & collision \\ 100 & reach \end{cases} \quad (4)$$

Therefore, an improved algorithm based on a dynamic reward function is proposed so that the mobile robot will not reduce its speed due to approaching obstacles during driving. This article applies the relationship between speed and heading angle to the setting of the reward function, and obtains different reward values by selecting different actions, as shown in Equation (5):

$$R_{couple} = \begin{cases} -100 & if\ hit \\ 20*(1+e^{-\frac{(v-\bar{v})^2}{2}}) & elif\ angle > 30 \\ 10*(1+e^{-\frac{(v-\bar{v})^2}{2}}) & elif\ angle \leq 30 \\ 10 & else \end{cases} \quad (5)$$

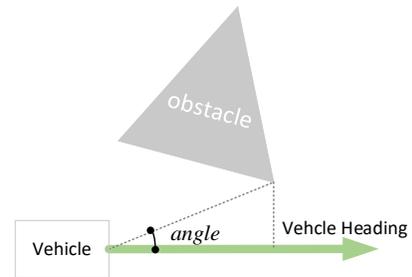

Figure 1. Obstacle angle determination

As shown in Fig.1, the *angle* is determined based on the current driving direction of the autonomous vehicle and the edge position of the obstacle. This can ensure that the vehicle will not be interfered by positional factors on the vehicle's speed planning when driving. If the vehicle's current heading is in line with the obstacle, If the distance is far apart, even if the vehicle is near an obstacle, it should maintain normal speed without slowing down. When the obstacle is on the vehicle's course, it will slow down appropriately to maintain safety.

In (5), $v$ represents the current speed of the vehicle, and $\bar{v}$ represents the expected speed of the vehicle. By adding the Gaussian function to the reward function, tthat can dynamically change the reward value according to the real-time status of the vehicle. If the deviation between the vehicle's current heading angle and the obstacle angle is greater than 30 degrees, the normal vehicle speed should be maintained at this time. When the angle between the vehicle's heading angle and the obstacle is less than 30 degrees, it should not slow down for static obstacles, but just keep passing at a constant speed. By coupling the reward of the unmanned vehicle with the current speed of the unmanned vehicle, the unmanned vehicle can still maintain an appropriate speed when passing the obstacle at a safe angle.

## III. EXPERIMENTS

All experiments are conducted on a computer equipped with Intel(R)Core(TM)i7-7700HQ CPU@2.80GHz and NVIDIA GeForceGTX1080GPU. All experiments are based on the ubuntu20.04 operating system and ROS, and the physical simulation is performed in Gazebo.

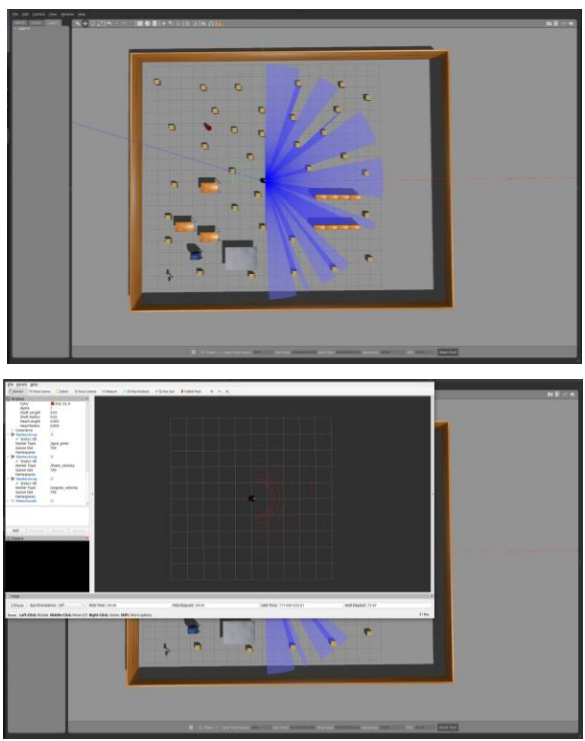

Figure 2. Simulation experiment environment

The experimental environment is shown in Fig.2. We constructed a 10*15 meter closed space and placed many obstacles in it to train DDQN.

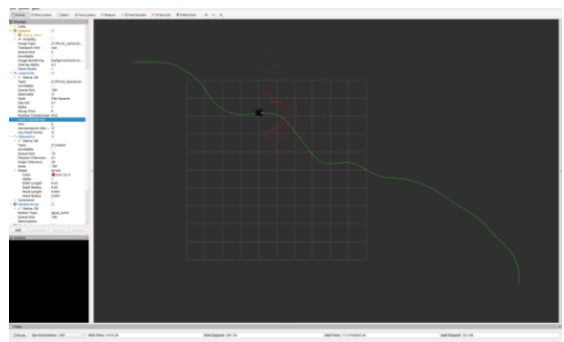

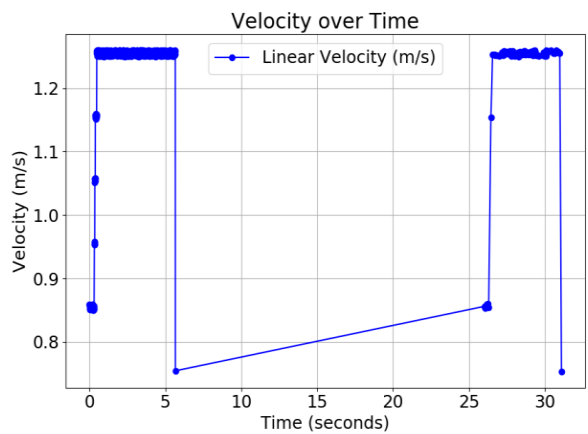

Figure 3. Driving trajectory and speed

Fig.3 shows the effect of a single plan in the simulation environment. As can be seen from the figure, although there are many obstacles in the environment, the average speed of unmanned vehicles can still reach more than 1.0. This algorithm can achieve relatively good results in environments with many obstacles.

TABLE I. AVERAGE SPEED IN DIFFERENT ENVIRONMENTS

|  | 10*15 METERS | 25*25 METERS |
|---|---|---|
| $R$ | 0.43(m/s) | 0.64(m/s) |
| $R_{couple}$ | 1.16(m/s) | 1.37(m/s) |

Table 1 shows the average driving speed of different reward functions in different environments. We conducted 20 experiments in each environment. It can be seen that, whether in a 10*15 environment or a 25*25 environment, the reward functions with coupling relationships perform well in speed planning. better. We set the desired speed to 1.2 meters per second. It can be seen from the table that the speed planning generated by the ordinary reward function has a low average speed, while the reward function with coupling relationship proposed in this article can make the speed reach the expected value without affecting the success rate of the plan.

## V. Conclusions

This article mainly studies the improvement of vehicle speed control by coupling the reward function and vehicle speed to each other. This article shows that using DDQN models and improved reward function can improve speed planning for autonomous vehicles. By updating the way the system responds to obstacles, the vehicle is able to maintain a steady speed without slowing down unnecessarily. Finally, the algorithm is simulated and verified by experiments in Gazebo environment. The results show that the improved reward function can carry out the corresponding speed planning according to the local environment and the number of obstacles, and can plan the traveling speed to meet the speed requirements under the conditions. Tests conducted in simulations confirmed that these methods work well in environments with various obstacles. The improvements to the reward function in this article can make unmanned vehicles more reliable and efficient in real-world environments.